\documentclass{article}

\usepackage{arxiv}

\usepackage[utf8]{inputenc} 
\usepackage[T1]{fontenc}    
\usepackage{hyperref}       
\usepackage{url}            
\usepackage{booktabs}       
\usepackage{amsfonts}       
\usepackage{nicefrac}       
\usepackage{microtype}      
\usepackage{natbib}
\usepackage{graphicx}
\usepackage[
  separate-uncertainty = true,
  multi-part-units = repeat
]{siunitx}
\usepackage{subcaption}
\usepackage{csvsimple}
\usepackage{amsmath}
\usepackage{cleveref}

\title{Weight Map Layer for Noise and Adversarial Attack Robustness}

\author{
  Mohammed Amer\thanks{Corresponding author} \\
  School of Computer Science\\
  University of Nottingham\\
  Semenyih, Malaysia \\
  \texttt{hcxma1@nottingham.edu.my} \\
   \And
  Tom\'as Maul\\
  School of Computer Science\\
  University of Nottingham\\
  Semenyih, Malaysia \\
  \texttt{tomas.maul@nottingham.edu.my} \\
}

\begin{document}
\maketitle

\begin{abstract}

Convolutional neural networks (CNNs) are known for their good performance and generalization in vision-related tasks and have become state-of-the-art in both application and research-based domains. However, just like other neural network models, they suffer from a susceptibility to noise and adversarial attacks. An adversarial defence aims at reducing a neural network's susceptibility to adversarial attacks through learning or architectural modifications. We propose the weight map layer (WM) as a generic architectural addition to CNNs and show that it can increase their robustness to noise and adversarial attacks. We further explain that the enhanced robustness of the two WM variants results from the adaptive activation-variance amplification exhibited by the layer. We show that the WM layer can be integrated into scaled up models to increase their noise and adversarial attack robustness, while achieving comparable accuracy levels across different datasets.

\end{abstract}

\section{Introduction}

Despite their wide adoption in vision tasks and practical applications, convolutional neural networks (CNNs) \citep{fukushima1980,LeCun1989, Krizhevsky2012} suffer from the same noise susceptibility problems manifested in the majority of neural network models. Noise is an integral component of any input signal that can arise from different sources, from sensors and data acquisition to data preparation and pre-processing.  \citet{Szegedy2013} opened the door to an extreme set of procedures that can manipulate this susceptibility by applying an engineered noise to confuse a neural network to misclassify its inputs.

The core principle in this set of techniques, called adversarial attacks, is to apply the least possible noise perturbation to the neural network input, such that the noisy input is not visually distinguishable from the original and yet it still disrupts the neural network output. Generally, adversarial attacks are composed of two main steps:

\begin{itemize}
    \item \textbf{Direction sensitivity estimation:} In this step, the attacker estimates which directions in the input are the most sensitive to perturbation. In other words, the attacker finds which input features will cause the most degradation of the network performance when perturbed. The gradient of the loss with respect to the input can be used as a proxy of this estimate.
    \item \textbf{Perturbation selection:} Based on the sensitivity estimate, some perturbation is selected to balance the two competing objectives of being minimal and yet making the most disruption to the network output.
\end{itemize}

The above general technique implies having access to the attacked model and thus is termed a whitebox attack. Blackbox attacks on the other hand assume no access to the target model and usually entail training a substitute model to approximate the target model and then applying the usual whitebox attack \citep{Chakraborty2018}. The effectiveness of this approach mainly depends on the assumption of the transferability between machine learning models \citep{Papernot2016b}.

Since their introduction, a lot of research have been done to guard against these attacks. An adversarial defence is any technique that is aimed at reducing the effect of adversarial attacks on neural networks. This can be through detection, modification to the learning process, architectural modifications or a combination of these techniques. Our approach consists of an architectural modification that aims to be easily integrated into any existing convolutional neural network architecture.

The core hypothesis we base our approach on starts from the premise that the noise in an input is unavoidable and in practise is very difficult to separate from the signal effectively. Instead, if the network can adaptively amplify the features activation variance selectively in its representations based on their importance, then it can absorb the variation introduced by the noise and map the representations to the correct output. This means that if a feature is very important to the output calculation, then its activation and intrinsic noise should be adequately amplified at training time to allow the classification layers to be robust to this feature's noisiness at inference time, since it is crucial to performance. In the context of CNNs, this kind of feature-wise amplification can be achieved by an adaptive elementwise scaling of feature maps.

We introduce the weight map layer (WM), which is an easy to implement layer composed of two main operations: elementwise scaling of feature maps by a learned weight grid of the same size, followed by a non-adaptive convolution reduction operation. We use two related operations in the two WM variants we introduce. The first variant, smoothing WM, uses a non-adaptive smoothing convolution filter of ones. The other variant, unsharp WM, adds an extra step to exploit the smoothed intermediate output of the first variant to implement an operation similar to unsharp mask filtering \citep{gonzalez2002}. The motivation for the second variant was to decrease the over-smoothing effect produced by stacking multiple WM layers. Smoothing is known to reduce adversarial susceptibility \citep{Xu2017}, however if done excessively this can negatively impact accuracy, which motivates the unsharp operation as a counter-measure to help control the trade-off between noise robustness and overall accuracy. We show and argue that the weight map component, can increase robustness to noise by amplifying the noise during the training phase in an adaptive way based on feature importance and, hence, can help networks absorb noise more effectively. In a way, this can be thought of as implicit adversarial training \citep{Goodfellow2014, Lyu2015, Shaham2015}. We show that the two components, weight map and reduction operations, can give rise to robust CNNs that are resistant to uniform and adversarial noise.

\section{Related Work}

Since the intriguing discovery by \citet{Szegedy2013} that neural networks can be easily forced to misclassify their input by applying an imperceptible perturbation, many attempts have been made to fortify them against such attacks. These techniques are generally applied to either learning or architectural aspects of networks. Learning techniques modify the learning process to make the learned model resistant to adversarial attacks, and are usually architecture agnostic. Architectural techniques, on the other hand, make modifications to the architecture or use a specific form of architecture engineered to exhibit robustness to such attacks.

\citet{Goodfellow2014} suggested adversarial training, where the neural network model is exposed to crafted adversarial examples during the training phase to allow the network to map adversarial examples to the right class. \citet{Tramer2017} showed that this can be bypassed by a two step-attack, where a random step is applied before perturbation. \citet{Jin2015} used a similar approach of training using noisy inputs, with some modifications to network operators to increase robustness to adversarial attacks. \citet{Seltzer2013} also applied a similar technique in the audio domain, namely, multi-condition speech, where the network is trained on samples with different noise levels. They also benchmarked against training on pre-processed noise-suppressed features and noise-aware training, a technique where the input is augmented with noise estimates. 

Distillation \citep{Hinton2015} was proposed initially as a way of transferring knowledge from a larger teacher network to a smaller student network. One of the tricks used to make distillation feasible was the usage of softmax with a temperature hyperparameter. Training the teacher network with a higher temperature has the effect of producing softer targets that can be utilized for training the student network. \citet{Papernot2016a, Papernot2017} showed that distillation with a high temperature hyperparameter can render the network resistant to adversarial attacks.  Feature squeezing \citep{Xu2017} corresponds to another set of techniques that rely on desensitizing the model to input, e.g. through smoothing images, so that it is more robust to adversarial attacks. This, however, decreases the model's accuracy. \citet{Hosseini2017} proposed NULL labeling, where the neural network is trained to reject inputs that are suspected to be adversarials. 

\citet{Sinha2018} proposed using adversarial networks to train the target network using gradient reversal \citep{Ganin2015}. The adversarial network is trained to classify based on the loss derived gradient, so that the confusion between classes with similar gradients is decreased. \citet{Pontes-Filho2018} proposed bidirectional learning, where the network is trained as a classifier and a generator, with an associated adversarial network, in two different directions and found that it renders the trained classifier more robust to adversarial attacks. 

From the architectural family, \citet{Lamb2018} proposed inserting Denoising Autoencoders (DAEs) between hidden layers. They act as regularizers for different hidden layers, effectively correcting representations that deviate from the expected distribution. A related approach was proposed by \citet{Ghosh2018}, where a Variational Autoencoder (VAE) was used with a mixture of Gaussians prior. The adversarial examples could be detected at inference time based on their high reconstruction errors and could then be correctly reclassified by optimizing for the latent vector that minimized the reconstruction error with respect to the input. DeepCloak \citep{Gao2017} is another approach that accumulates the difference in activations between the adversarials and the seeds used to generate them at inference time and, based on this, a binary mask is inserted between hidden layers to zero out the features with the highest contribution to the adversarial problem. The nearest to our approach, is the method proposed by \citet{Sun2017}. This work made use of a HyperNetwork \citep{Ha2016} that receives the mean and standard deviation of the convolution layer and outputs a map that is multiplied elementwise with the convolution weights to produce the final weights used to filter the input. The dependency of the weights on the statistics of the data renders the network robust to adversarial attacks.

We introduce the WM layer, an adversarial defence which requires a minimal architectural modification since it can be inserted between normal convolutional layers. We propose that the adaptive activation-variance amplification achieved by the layer, which can be considered as a form of dynamic implicit adversarial training, can render CNNs robust to different forms of noise. Finally, we show that the WM layer can be integrated into scaled up models to achieve noise robustness with the same or similar accuracy in many cases across different datasets.

\section{Methods}

\begin{figure}
    \centering
    \includegraphics[width=0.75\textwidth]{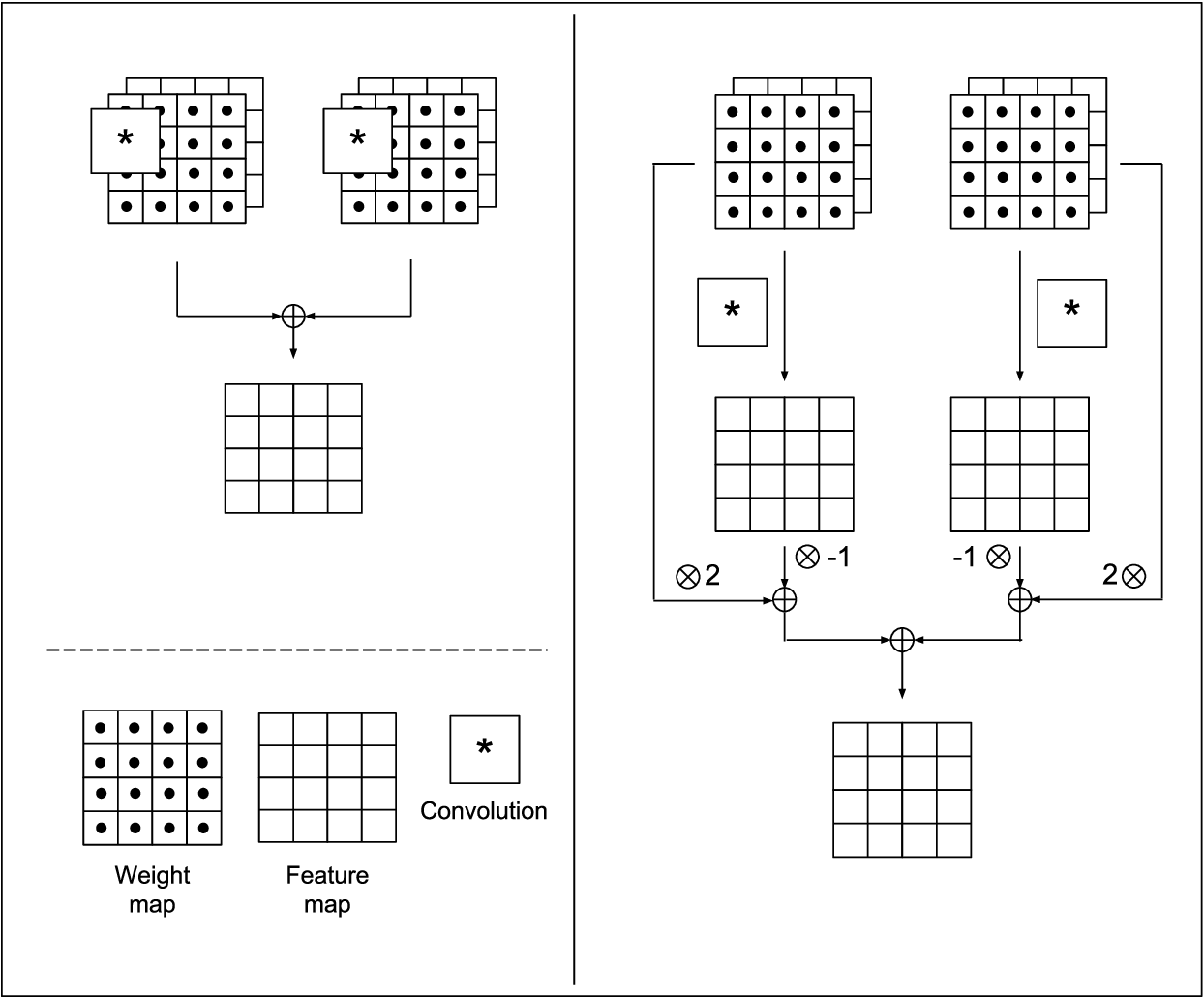}
    \caption{Weight map layer. Left: smoothing. Right: unsharp.}
    \label{fig:weight-map}
\end{figure}

The main operation involved in a weight map layer \cref{fig:weight-map} is an elementwise multiplication of the layer input with a map of weights. For a layer $l$ with an input $x_l \in R^{C_i \times D_i \times D_i}$ with $C_i$ input channels and spatial dimension $D_i$ and an output $o_l \in R^{C_o \times D_o \times D_o}$ with $C_o$ output channels and $D_o$ spatial dimension, the channel map of the $c_i$th input channel contributing to the $c_o$th output channel is calculated as

\begin{equation}
    m_l^{(c_i,c_o)} = W_l^{(c_i,c_o)} \odot x_l^{(c_i)}
\end{equation}

where $W_l^{(c_i,c_o)} \in R^{D_i \times D_i}$ is the weight mapping between $c_i$ and $c_o$, $x_l^{(c_i)}$ is the $c_i$th input channel and $\odot$ is the elementwise multiplication operator. We used two techniques for producing the pre-nonlinearity output of the weight map layer. The first variant, smoothing weight map layer, produces the $c_o$th output channel $o_l^{(c_o)}$ by convolving the maps with a kernel $k \in R^{C_i \times D_k \times D_k}$ of ones with $D_k$ spatial dimension as follow,

\begin{equation}
    o_l^{(c_o)} = m_l^{(c_o)} \ast k + b_l^{(c_o)}
\end{equation}

where $m_l^{(c_o)}$ is the set of intermediate maps contributing to output channel $c_o$, $b_l^{(c_o)} \in R^{D_o \times D_o}$ is a bias term and $\ast$ is the convolution operator. The other variant, unsharp weight map layer, produces the output by an operation similar to unsharp filtering as follow,

\begin{equation}
    s_l^{(c_i,c_o)} = 2 m_l^{(c_i,c_o)} - m_l^{(c_i,c_o)} \ast k
\end{equation}

\begin{equation}
    o_l^{(c_o)} = \sum_{c_i} s_l^{(c_i,c_o)} + b_l^{(c_o)}
\end{equation}

where $k \in R^{D_k \times D_k}$ is a kernel of ones applied with a suitable padding element to ensure similar spatial dimensions between the convolution input and output.

\section{Experiments}


\begin{table}
    \centering
    \begin{tabular}{|c|c|c|c|}
    \hline
    \textbf{Layer} & \textbf{CNN}& \textbf{CNN (wide)} & \textbf{WM} \\
    \hline
    1 & Conv(33 channels) & Conv(200 channels) & WM (32 channels)\\
    \hline
    2 & Conv(33 channels)  & Conv(500 channels)& WM (32 channels)\\
    \hline
    3 & Conv(8 channels) & Conv(8 channels) & WM (8 channels)\\
    \hline
    4 & FC(64 nodes) & FC(64 nodes) & FC(64 nodes)\\
    \hline
    5 & FC(10 nodes) & FC(10 nodes) & FC(10 nodes)\\
    \hline
    \end{tabular}
    \caption{CNN variants basic skeletons}
    \label{tab:cnn-skeleton}
\end{table}

\begin{table}
    \centering
    \begin{tabular}{|c|c|c|}
    \hline
    \textbf{Layer} & \textbf{Out dimension} & \textbf{Repeat} \\
    \hline
    ResBlock & 8 & 3 \\
    \hline
    ResBlock & 16 & 4 \\
    \hline
    ResBlock & 32 & 6 \\
    \hline
    ResBlock & 64 & 4 \\
    \hline
    Global average pool & 64 & 1\\
    \hline
    Fully connected & 10 & 1 \\
    \hline
    \end{tabular}
    \caption{ResNet skeleton}
    \label{tab:resnet-arch}
\end{table}

\begin{table}
    \centering
    \begin{tabular}{|c|c|c|}
    \hline
    \textbf{Layer} & \textbf{Hyperparams} & \textbf{Repeat} \\
    \hline
    Conv & channels: 16 & 1 \\
    \hline
    ReLU & channels: 16 & 1 \\
    \hline
    Dense & Growth rate: 8 & 2 \\
    \hline
    Max pool & size:2, stride:2 & 1 \\
    \hline
    Dense & Growth rate: 8 & 4 \\
    \hline
    Max pool & size:2, stride:2 & 1 \\
    \hline
    Dense & Growth rate: 8 & 8 \\
    \hline
    Max pool & size:2, stride:2 & 1 \\
    \hline
    Dense & Growth rate: 8 & 16 \\
    \hline
    Global average pool & out: 256 & 1\\
    \hline
    Fully connected & out: 10 & 1\\
    \hline
    \end{tabular}
    \caption{DenseNet skeleton}
    \label{tab:densenet-arch}
\end{table}

We tested our method on three architectures and three datasets. For every dataset, we benchmark a number of weight-map layer variants against a baseline. The baseline and the variants share the same skeleton, but the layers in the baseline are convolutional layers, while in the different weight map variants some or all of the convolutional layers are replaced with a variant of the weight map layer. We train each model for 300 epochs using the Adam optimizer. The test error reported on the non-noisy dataset is an average of three trials. For the weight map layer, we initialize the weights and biases according to

\begin{align*}
    w_{ij} \sim \emph{U}(-s, s)\\
    s = \frac{1}{\sqrt{a_{in}}}
\end{align*}

where $\emph{U}$ is the normal distribution and $a_{in}$ is the fan-in.

In the preliminary experiments on the MNIST dataset, we tested three different architectures: CNN, ResNet \citep{He2015} and DenseNet \citep{Huang2016} and for the two other datasets (CIFAR10 and iWild Cam2019) we tested ResNet and DenseNet. The CNN skeleton is a stack of three layers, where the first two have either 32 channels, if it is a weight map network variant, or 33 channels if it is a normal CNN \cref{tab:cnn-skeleton}. This difference was adopted to maintain approximately the same number of floating point operations per second (FLOPS) between the two architectures. In just one of the CNN variants, we increased the channels of the first two layers to 200 and 500, respectively, to compare with the weight map network having the same number of parameters. We will refer to this scaled up variant as "wide" in the results. The final layer in the skeleton body has 8 channels.  Classification output is made by a 2 layer fully connected multilayer perceptron (MLP), where the first layer has 64 nodes followed by an output layer. We fixed the kernel size across all the layers. We compare two kernel sizes, 3 and 9, and we include batchnorm \citep{Ioffe2015} layers in some of the variants to test the interaction with the proposed layer. When batchnorm is included, it is inserted in all the convolutional layers just before the nonlinearity. In one of the variants, we elementwise multiply the input with a learned weight map to probe the effect of the input weight map on noise robustness.

To assess the scalability of the proposed weight map layer, we integrated it into two popular CNN skeletons: ResNet and DenseNet. \Cref{tab:resnet-arch} shows the skeleton of the ResNet variant. ResBlock was composed of two layers of 3x3 convolutions with ReLU activations. At layer transitions characterized by doubling of the number of channels, downsampling to half of the spatial dimension was done by the first layer of the first block. Residual connections were established from the input to each ResBlock to its output, following the pattern used in the original paper \citep{He2015}, where projections using 1x1 convolutions were applied when there was a mismatch of the number of channels or spatial dimensions. \Cref{tab:densenet-arch} shows the skeleton of DenseNet. Each Dense layer is assumed to be followed by a ReLU nonlinearity. For integrating WM layers into the architectures, we either replace all the layers by one of the WM layer variants or replace half of the layers by skipping one layer and replacing the next. We will refer to the former by the non-alternating WM model and to the latter by the alternating WM model. 

We tested model robustness on two types of noise, uniform noise and adversarial noise. When testing models for uniform noise robustness, we added random uniform noise to the input, which always had a lower boundary of zero. We varied the upper boundary to assess the degree of robustness. After the addition of the noise, the input was renormalized to be within the range $[0, 1]$. The robustness measure is reported as the average test error achieved by the model on the noisy test dataset averaged over three trials. For testing the models against adversarial attacks, we followed the fast gradient sign method (FSGM) \citep{Goodfellow2014} approach, where we varied the $\epsilon$ parameter to control the severity of the attack.

\subsection{MNIST}

The MNIST dataset is a collection of images of handwritten digits with 10 classes representing the digits from 0 to 9. The images are grayscale with spatial dimensions of 28x28. MNIST consists of a training dataset of 60k samples and a testing dataset of 10k samples. We divide the training dataset into 90\% for training and 10\% for validation. The only augmentation used during the training was padding by 2 and then random cropping a 28x28 patch. The test dataset was used as it is. The results on MNIST are summarized in \cref{tab:mnist-res} and \cref{fig:mnist-noise-res}.

\subsection{CIFAR10}

The CIFAR10 dataset is a collection of colored (3-channels) images comprising 10 classes representing animals and vehicles. The images have 32x32 spatial dimensions. CIFAR10 consists of 50k training images and 10k testing images. We divide the training set into 90\% for training and 10\% for validation. No augmentation was used for the training data. The test dataset was used as it is. CIFAR10 results are summarized in \cref{tab:cifar10-res} and \cref{fig:cifar-noise-res}.

\subsection{iWildCam2019}

The iWildCam2019 dataset is a collection of colored animal images captured in the wild by camera traps. The dataset is imbalanced and there is only a partial overlap between the classes found in the training and test datasets. To balance the dataset and fix the classes between training and testing, we used only the training set by splitting it into training, validation and test sets. This was done by first choosing 10 classes (deer, squirrel, rodent, fox, coyote, raccoon, skunk, cat, dog, opossum), then balancing all the classes by choosing only 1000 samples from each class. We then split the data into 70\% training, 20\% validation and 10\% test. This was done class-wise to maintain the balance. We preprocessed all the sets by converting the images to grayscale and downsampling to 23x32 spatial dimensions. No augmentation was used. iWildCam2019 results are summarized in \cref{tab:iwild-res} and \cref{fig:iwild-noise-res}.

\subsection{Results}


\begin{table}
    \centering
    \begin{tabular}{|c|c|c|c|c|}
    \hline
    \textbf{Arch} & \textbf{Variant} & \textbf{Params} & \textbf{GFLOPS} & \textbf{Test error} (\%) \\
    \hline
    CNN & wide & 1.19M & 1.07 & \SI{1.0 \pm 0.09}{} \\
    \hline
    CNN & 33 channels & 261K & 0.014 & \SI{0.85 \pm 0.11}{} \\
    \hline
    CNN & 33 channels - batchnorm & 261K & 0.014 & \SI{0.73 \pm 0.10}{} \\
    \hline
    CNN & 33 channels - kernel size 9 & 361K & 0.125 & \SI{0.7 \pm 0.09}{} \\
    \hline
    CNN & 33 channels - input-scale & 261K & 0.014 & \SI{0.86 \pm 0.01}{} \\
    \hline
    Smooth WM & 32 channels & 1.16M & 0.013 & \SI{0.79 \pm 0.03}{} \\
    \hline
    Smooth WM & 32 channels - batchnorm & 1.16M & 0.013 & \SI{0.7 \pm 0.07}{} \\
    \hline
    Smooth WM & 32 channels - kernel size 9 & 1.16M & 0.119 & \SI{0.88 \pm 0.04}{} \\
    \hline
    Unsharp WM & 32 channels & 1.49M & 0.020 & \SI{0.73 \pm 0.03}{} \\
    \hline
    \end{tabular}
    \caption{CNN results (MNIST)}
    \label{tab:cnn-res}
\end{table}

\begin{table}
    \centering
    \begin{tabular}{|c|c|}
    \hline
    \textbf{Variant} & \textbf{Test error} (\%) \\
    \hline
    \multicolumn{2}{|l|}{\textbf{ResNet}}\\
    \hline
    Conv & \SI{0.5 \pm 0.05}{}\\
    \hline
    Smoothing WM & \SI{0.8 \pm 0.09}{}\\
    \hline
    Unsharp WM & \SI{0.91 \pm 0.14}{}\\
    \hline
    Alternating Conv/Smoothing WM & \SI{0.65 \pm 0.08}{}\\
    \hline
    Alternating Conv/Unsharp WM & \SI{0.71 \pm 0.1}{}\\
    \hline
    \multicolumn{2}{|l|}{\textbf{DenseNet}}\\
    \hline
    Conv & \SI{0.52 \pm 0.09}{}\\
    \hline
    Smoothing WM & \SI{0.67 \pm 0.05}{}\\
    \hline
    Unsharp WM & \SI{0.6 \pm 0.04}{}\\
    \hline
    Alternating Conv/Smoothing WM & \SI{0.55 \pm 0.07}{}\\
    \hline
    Alternating Conv/Unsharp WM & \SI{0.54 \pm 0.04}{}\\
    \hline
    \end{tabular}
    \caption{MNIST results}
    \label{tab:mnist-res}
\end{table}

\begin{table}
    \centering
    \begin{tabular}{|c|c|}
    \hline
    \textbf{Variant} & \textbf{Test error} (\%) \\
    \hline
    \multicolumn{2}{|l|}{\textbf{ResNet}}\\
    \hline
    Conv & \SI{32.14 \pm 2.2}{}\\
    \hline
    Alternating Conv/Smoothing WM & \SI{39.31 \pm 0.7}{}\\
    \hline
    Alternating Conv/Unsharp WM & \SI{38.49 \pm 0.9}{}\\
    \hline
    \multicolumn{2}{|l|}{\textbf{DenseNet}}\\
    \hline
    Conv & \SI{23.18 \pm 0.7}{}\\
    \hline
    Alternating Conv/Smoothing WM & \SI{29.04 \pm 0.4}{}\\
    \hline
    Alternating Conv/Unsharp WM & \SI{30.15 \pm 0.8}{}\\
    \hline
    \end{tabular}
    \caption{CIFAR10 results}
    \label{tab:cifar10-res}
\end{table}

\begin{table}
    \centering
    \begin{tabular}{|c|c|}
    \hline
    \textbf{Variant} & \textbf{Test error} (\%) \\
    \hline
    \multicolumn{2}{|l|}{\textbf{ResNet}}\\
    \hline
    Conv & \SI{23.3 \pm 1.0}{}\\
    \hline
    Alternating Conv/Smoothing WM & \SI{20.43 \pm 0.8}{}\\
    \hline
    Alternating Conv/Unsharp WM & \SI{20.4 \pm 0.4}{}\\
    \hline
    \multicolumn{2}{|l|}{\textbf{DenseNet}}\\
    \hline
    Conv & \SI{22.53 \pm 0.8}{}\\
    \hline
    Alternating Conv/Smoothing WM & \SI{22.47 \pm 0.7}{}\\
    \hline
    Alternating Conv/Unsharp WM & \SI{20.7 \pm 1.0}{}\\
    \hline
    \end{tabular}
    \caption{iWildCam2019 results}
    \label{tab:iwild-res}
\end{table}

The test errors of the preliminary experiments benchmarking CNN on MNIST are summarized in \cref{tab:cnn-res}. The basic weight map network has better performance than the corresponding basic CNN with the same number of FLOPS. The unsharp version is better by a larger margin but with slightly higher FLOPS. Increasing the CNN parameters to the level of the corresponding weight map network results in lowering its performance. Including batchnorm in either the CNN or the weight map network boosted the performance of both variants to nearly the same level. On the other hand, increasing the kernel size to 9 boosted the CNN performance, whilst degrading the weight map network performance. 

The test error results of ResNet and DenseNet are summarized in \cref{tab:mnist-res}, \cref{tab:cifar10-res} and \cref{tab:iwild-res}. 
Baseline ResNet had better performance than the weight map layer variants on MNIST and CIFAR10, while all of the weight map layer variants were better than the baseline on iWildCam2019. For DenseNet tested on MNIST, the baseline had better performance than the non-alternating weight map layer variants, while it had similar performance to the alternating variants. On iWildCam2019, the baseline DenseNet had similar performance to the alternating smoothing variant and lower performance than the alternating unsharp variant. On CIFAR10, the weight map variants had lower performance than the baseline DenseNet.

\begin{figure}

    \begin{subfigure}[t]{\textwidth}
        \centering
        \includegraphics[width=\textwidth]{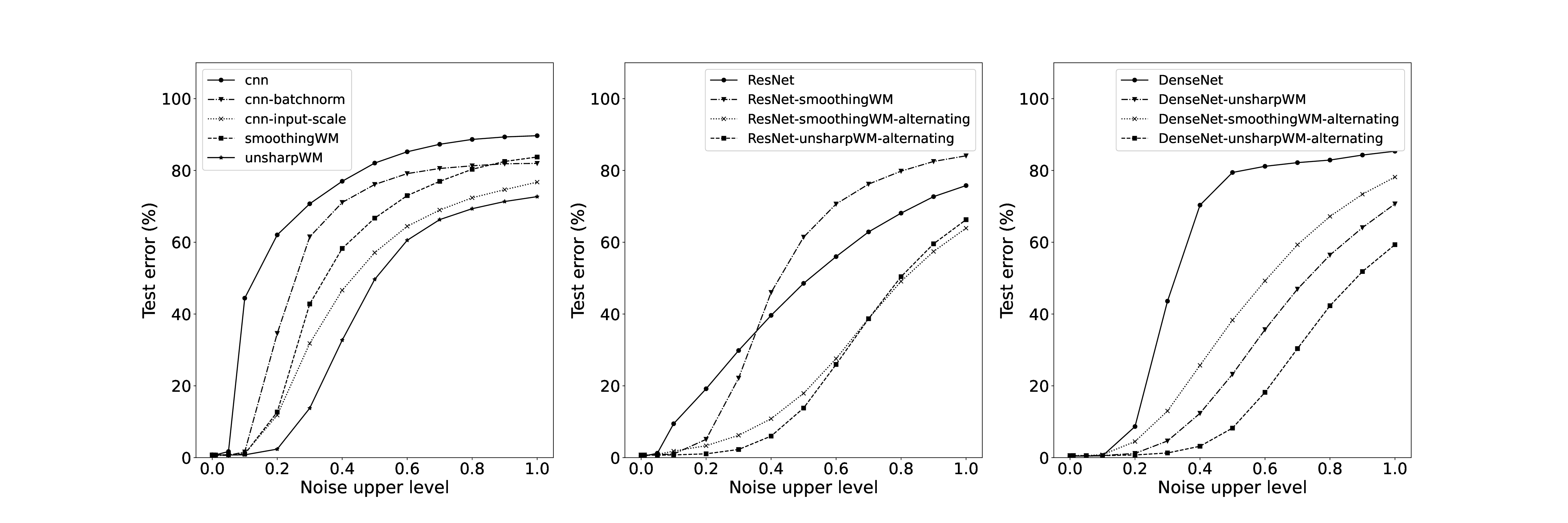}
        \caption{Uniform noise}
    \end{subfigure}
    
    \begin{subfigure}[b]{\textwidth}
        \centering
        \includegraphics[width=\textwidth]{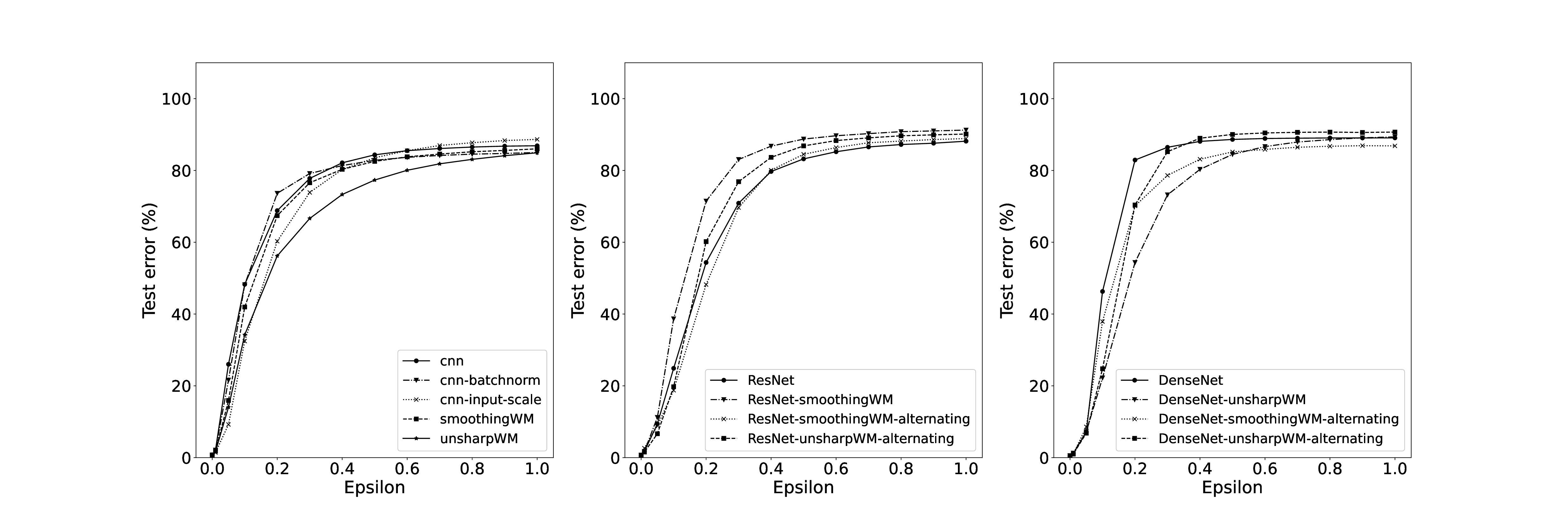}
        \caption{Adversarial noise}
    \end{subfigure}
    \caption{MNIST. Left: CNN variants. Middle: ResNet variants. Right: DenseNet variants.}
    \label{fig:mnist-noise-res}
\end{figure}

\begin{figure}

    \begin{subfigure}[t]{\textwidth}
        \centering
        \includegraphics[width=\textwidth]{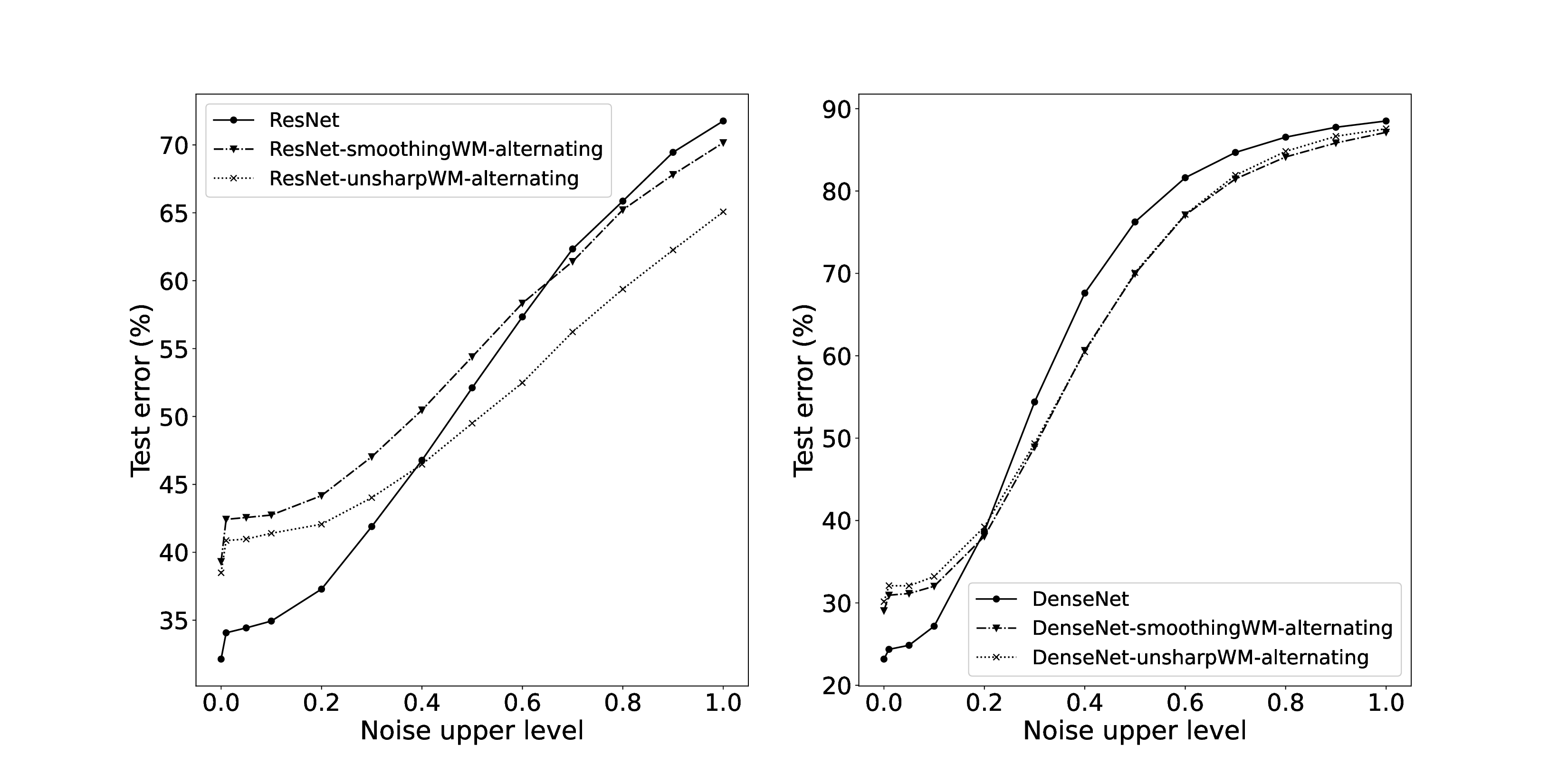}
        \caption{Uniform noise}
    \end{subfigure}
    
    \begin{subfigure}[b]{\textwidth}
        \centering
        \includegraphics[width=\textwidth]{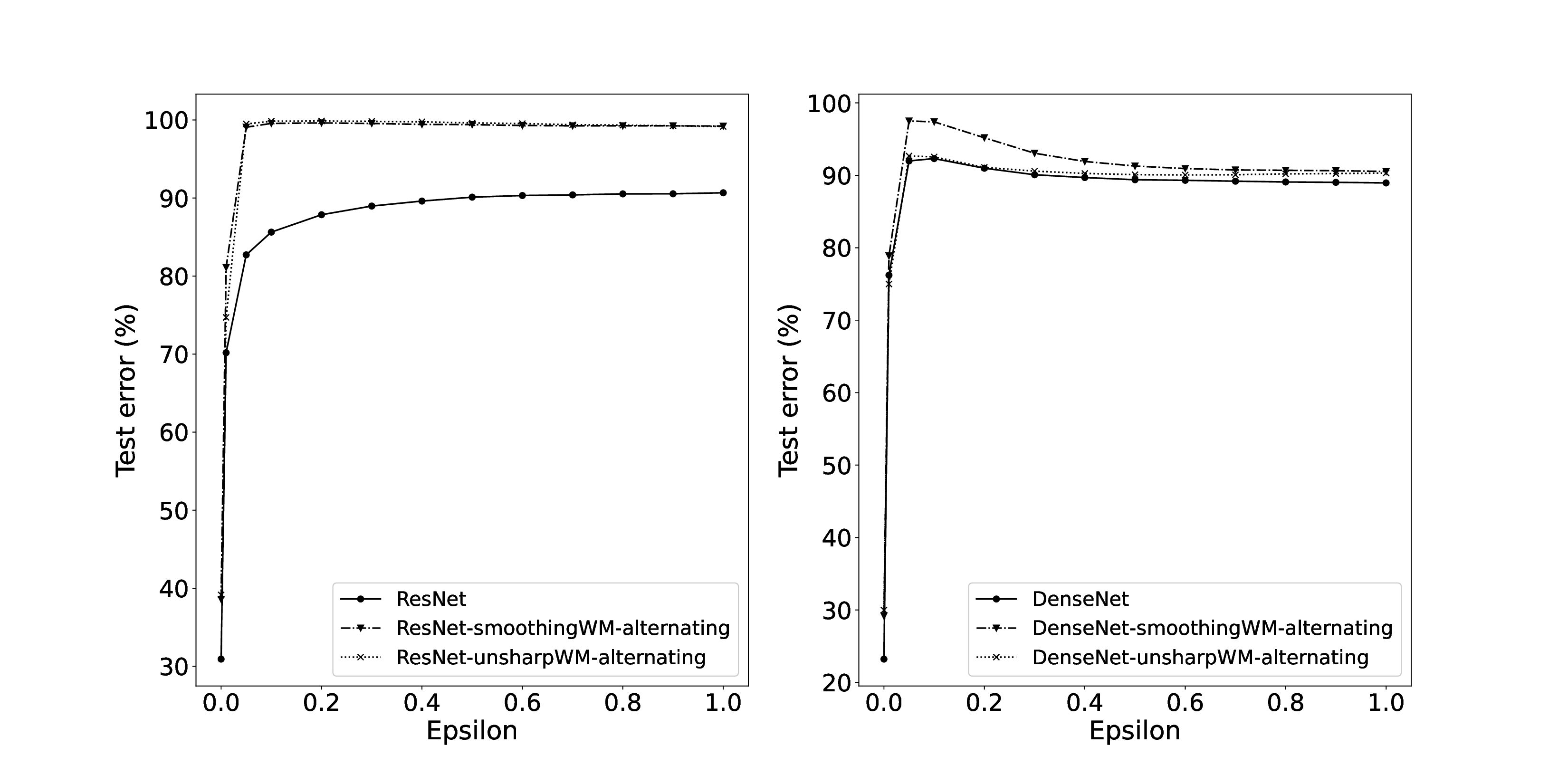}
        \caption{Adversarial noise}
    \end{subfigure}
    \caption{CIFAR10. Left: ResNet variants. Right: DenseNet variants.}
    \label{fig:cifar-noise-res}
\end{figure}

\begin{figure}

    \begin{subfigure}[t]{\textwidth}
        \centering
        \includegraphics[width=\textwidth]{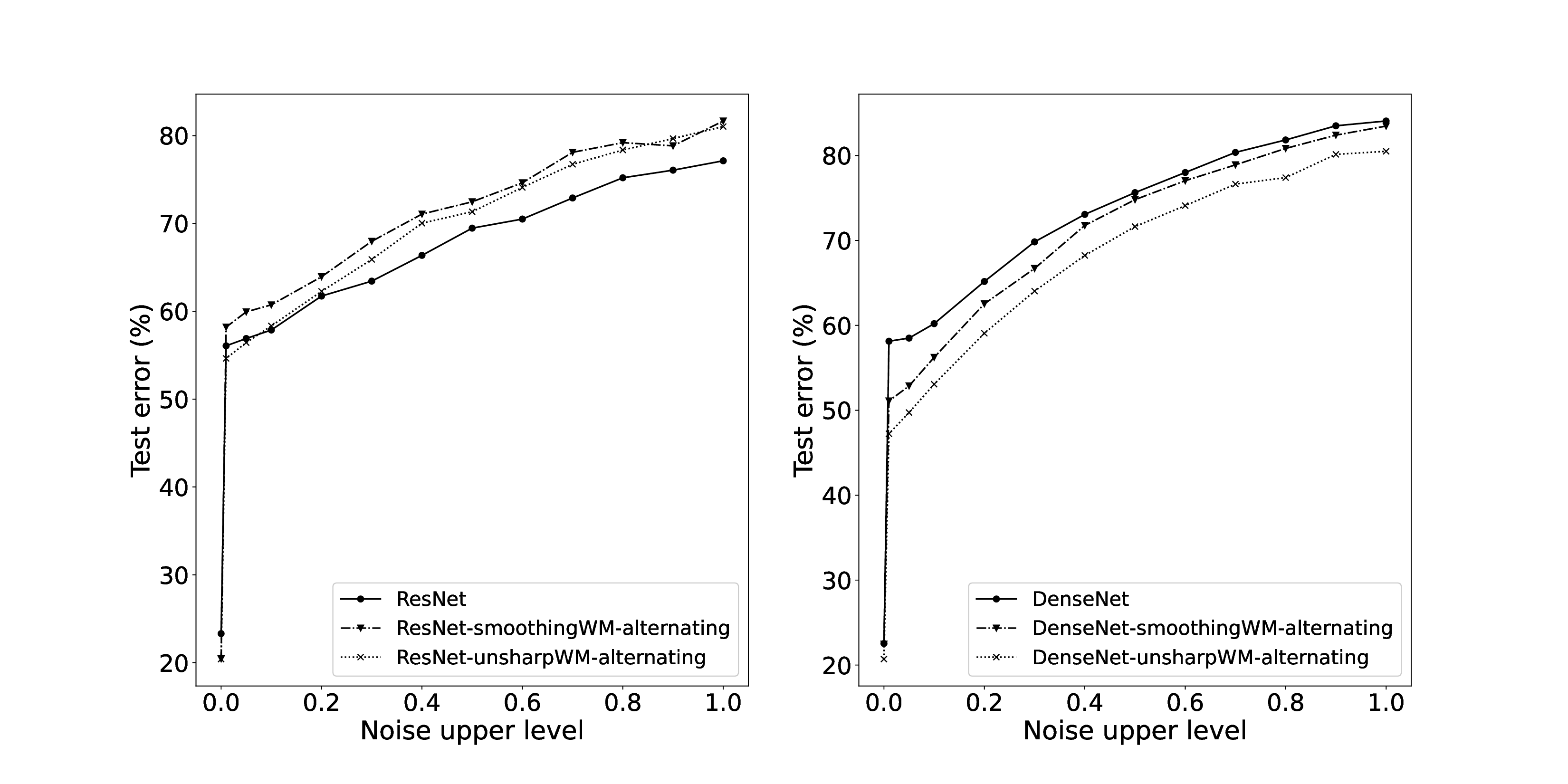}
        \caption{Uniform noise}
    \end{subfigure}
    
    \begin{subfigure}[b]{\textwidth}
        \centering
        \includegraphics[width=\textwidth]{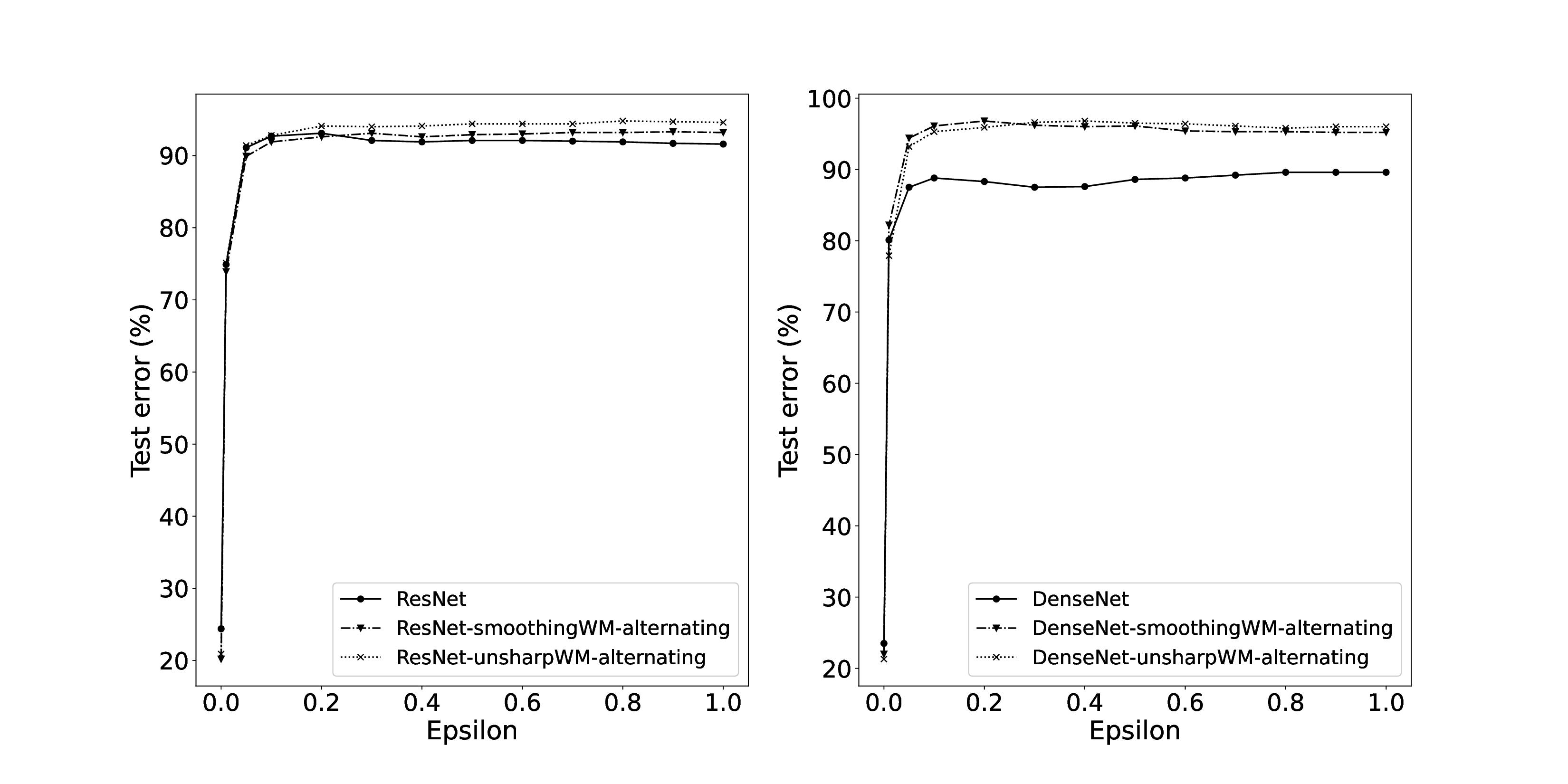}
        \caption{Adversarial noise}
    \end{subfigure}
    \caption{iWildCam2019. Left: ResNet variants. Right: DenseNet variants.}
    \label{fig:iwild-noise-res}
\end{figure}

Robustness to noise results are shown in \cref{fig:mnist-noise-res}, \cref{fig:cifar-noise-res} and \cref{fig:iwild-noise-res}. For uniform noise, all of the WM variants tested on MNIST had better performance than the baseline, except for the non-alternating smoothing WM variant of ResNet. On CIFAR10, the alternating unsharp WM variant of ResNet showed lower performance than the baseline at lower noise, however, it becomes more robust than the baseline as the noise magnitude increases. All the alternating WM variants of DenseNet showed a similar behaviour. The alternating smoothing WM variants of ResNet, however, had less robustness at all noise levels. On iWildCam2019, all of the WM variants of ResNet were less robust than the baseline, while all the WM variants of DenseNet were more robust than the baseline. 

Regarding adversarial noise results on MNIST, the unsharp WM variant of CNN was more robust than the baseline, while the baseline and all the other WM variants of CNN had similar performance. For ResNet, all the WM variants had either less or the same robustness compared to the baseline. All the WM variants of DenseNet were better than the baseline, except the alternating unsharp WM model which had similar performance to the baseline. On CIFAR10, both the baselines and the WM models of ResNet performed poorly, however, on average baseline was better than WM variants of ResNet, while WM variants of DenseNet were better on average than their baseline. iWildCam2019 had similar results to CIFAR10, where all the models, including the baseline, performed poorly. All ResNet models on iWildCam2019 had similar performance, while for DenseNet models, the baseline was on average better than the WM variants. 

\section{Discussion}

We have benchmarked two reduction operations in WM layers: smoothing and unsharp. The smoothing operation is effectively a moving average over a window equal to the kernel area. Since this will introduce blurriness into the input which will accumulate further on stacking
multiple layers, we benchmarked against another reduction operation that mimics the action of an unsharp filter to reduce the accumulating blurring effect. To further reduce the introduced distortion, we benchmarked models which alternate between having a normal convolutional layer and a WM layer.

\begin{figure}
    \centering
    \includegraphics[width=\textwidth]{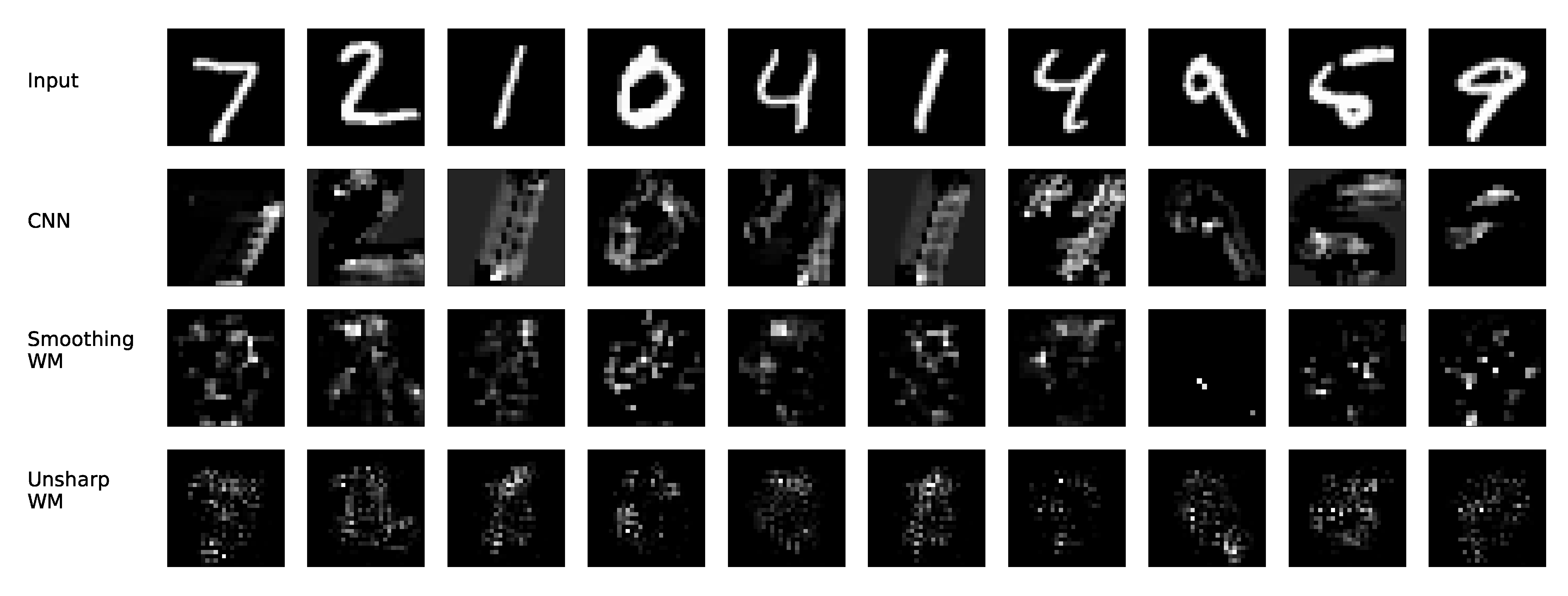}
    \caption{Grad-CAM visualization}
    \label{fig:prelim-cam}
\end{figure}

For the preliminary experiments based on CNN, the WM variant (no batchnorm and kernel size of 3) has a better performance than the corresponding vanilla CNN having the same number of FLOPS. We attribute this to two main factors. First, the higher capacity of the WM variant, due to its larger number of parameters, makes it more expressive. WM representation doesn't, however, need to be in the same space as the CNN variant. The Grad-CAM \citep{Selvaraju2016} visualization of both vanilla CNN and the two WM variants \cref{fig:prelim-cam} shows a substantial difference. While the CNN CAM is a blurry, diffused distortion of the input and sometimes activating for a large proportion of the background, the WM CAM is sharper, sparser and more localized with almost no diffused background activation, specially for the unsharp WM variant. We attribute this background activation sparsity to the feature selection ability of WM. Much like the way attentional techniques \citep{Bahdanau2014,Vinyals2014,Xu2015,Hermann2015} can draw the network to focus on a subset of features, WM includes an elementwise multiplication by a weight map, that can in principle achieve feature selection on a pixel by pixel basis. On the other hand, normal convolution can't achieve a similar effect because of weight sharing. The second possible reason for better performance consists of the scaling properties of the WM layer. This can in principle act like the normalization done by batchnorm layers. However, applying batchnorm can boost the performance of both CNN and WM variants, which indicates that the two approaches have an orthogonal component between them. Moreover, as we discuss below, batchnorm alone doesn't protect against uniform noise and adversarial attacks. If we fix the number of parameters, instead of FLOPS, along with depth, we observe a clear advantage for WM variants. The WM variant with the same number of parameters and depth is better in performance by a large margin, and cheaper in FLOPS by around 100x. We attribute this to the large width compared to depth of the CNN variant, which makes it harder to optimize. On the other hand, WM can pack larger degrees of freedom without growing in width. 

Increasing the kernel size enhances the performance of the CNN variant, while it lowers the performance of the WM variant. The enhanced CNN performance is due to increased capacity and a larger context made available by the larger receptive field. In the case of WM, the increased kernel size results in over smoothing and larger overlapping between adjacent receptive fields, effectively sharing more parameters and limiting the model's effective capacity.

For deeper networks like ResNet and DenseNet, we will focus the discussion on the alternating models. The unsharp models show a similar or better test performance compared to the smoothing models across all datasets. This supports our motivation for introducing the unsharp operation to reduce distortion.

For uniform noise, we find that for all the WM models in MNIST and CIFAR10, at least one of the WM variants have better noise robustness than the baseline. In this subset, all of the WM models of CNN and DenseNet in MNIST have better noise robustness than the baseline. For uniform noise in iWildCam2019, the WM variants of DenseNet are more robust than the baseline, while the WM variants of ResNet are less robust than the baseline. For adversarial attacks, the WM variants of CNN and DenseNet have at least one model which is better than the baseline. For ResNet on MNIST, WM variants have the same as or less robustness than the baseline. For CIFAR10 and iWildCam2019, all the models are very sensitive to the adversarial noise, with mixed relative results between the WM variants and the baseline.

The above summary of results shows that the WM layer is very effective against random noise, while it doesn't have the same efficacy against adversarial attacks. This may be explained by our hypothesis on the mechanism of the WM layer. The elementwise multiplication of the weight map by the input pixels, or any intermediate activation in general, can be considered as a feature selection mechanism. This implies that pixels that have large contribution to the accuracy will have large magnitude. The variance of any activation after applying the elementwise multiplication is then

\begin{equation}
    \mathrm{Var}[w_{ij} \odot x_{ij}] = w^2_{ij} \odot \mathrm{Var}[x_{ij}]
\end{equation}

where $x_{ij}$ is the activation at the index $ij$, $\mathrm{Var}$ is the variance of that activation over the dataset, $w_{ij}$ is the weight corresponding to the activation and $\odot$ is the elementwise multiplication operator.

We attribute the noise resistance introduced by the WM layer to the described amplification of the activation variance. This amplification is proportional to the weight magnitude and, hence, to the importance of the corresponding feature to the model performance. The amplification of variance increases the variations that the network encounters for the corresponding pixel/activation, which can be considered as an implicit augmentation of the dataset. This hypothesis, in turn, explains why it is performing better on random noise than adversarial attacks. While random noise has arbitrary direction, an adversarial attack can exploit gradient information to account for any differentiable component.  

\section{Conclusion}

We introduced the weight map layer with its two variants as a generic architectural modification that can increase the robustness of convolutional neural networks to noise and adversarial attacks. We showed that it can be used to boost performance and increase noise robustness in small convolutional networks. Moreover, we showed that WM layers can be integrated into scaled up networks, ResNet and DenseNet, to increase their noise and adversarial attack robustness, while achieving comparable accuracy. We explained that the adaptive activation-variance amplification exhibited by the WM layer can explain its noise and adversarial attack robustness and the associated experimental observations regarding its dynamics. Future work has multiple promising directions with regards to finding more effective ways to integrate with more architectures to achieve noise robustness with more accuracy, providing more insights and experimental results regarding the dynamics of the WM layer and exploiting this further to enhance the accuracy and noise robustness of neural networks in general.

\section{Acknowledgement}

This work was partially supported by a grant from Microsoft's AI for Earth program.

\bibliographystyle{plainnat}
\bibliography{main}

\end{document}